\documentstyle[aaai,lingmacros,psfig]{article} 

\begin{document}

\title{Improvising Linguistic Style: Social and Affective Bases for
Agent Personality}

\author{\begin{tabular}{ccc}
Marilyn A. Walker   & Janet E. Cahn & Stephen J. Whittaker \\ 
AT\&T Labs Research & MIT Media Lab & AT\&T Labs Research 
\thanks
{This research was supported in part by Hewlett Packard
Laboratories U.K., by the University of Pennsylvania, by Mitsubishi
Electric Research Laboratories, by the MIT Media Laboratory, and by
AT\&T Research Laboratories.} 
\thanks
{Permission to copy without fee all or or part of this material is
granted provided that the copies are not made or distributed for
direct commercial advantage, the ACM copyright notice and the title
of the publication and its date appear, and notice is given that
copying is by permission of ACM. To copy otherwise, or to
republish, requires a fee and/or specific permission. Agents '97
Conference Proceedings, copyright 1997 ACM.} 
\\
600 Mountain Ave.     & 20 Ames St.          & 600 Mountain Ave.  \\
Murray Hill, NJ 07974 & Cambridge, MA  02139 & Murray Hill, NJ 07974\\
{\sf walker@research.att.com} & {\sf  cahn@media.mit.edu} 
& {\sf  whittaker@research.att.com} 
\end{tabular}
}
\maketitle

\begin{abstract} This paper introduces Linguistic Style
Improvisation, a theory and set of algorithms for improvisation of
spoken utterances by artificial agents, with applications to
interactive story and dialogue systems. We argue that linguistic
style is a key aspect of character, and show how speech act
representations common in AI can provide abstract representations
from which computer characters can improvise. We show that the
mechanisms proposed introduce the possibility of socially oriented
agents, meet the requirements that lifelike characters be
believable, and satisfy particular criteria for improvisation
proposed by Hayes-Roth. \end{abstract}

\section{Introduction}
\label{intro-sec}

\begin{quote} {\small {\sf Just because you are a character doesn't
mean that you have character.} {\footnotesize Wolf to Raquel in
{\em Pulp Fiction}, Q. Tarantino.}} \end{quote}

Linguistic Style Improvisation (henceforth LSI) concerns the
choices that speakers make about the {\sc semantic content}, {\sc
syntactic form} and {\sc acoustical realization} of their spoken
utterances. This paper argues that linguistic style is a key aspect
of an agent's character. We present a novel theory of, and
algorithms for, Linguistic Style Improvisation by computer
characters.

As an example of how linguistic style can convey character,
consider Victor Laszlo's request for two cointreaux in \ex{1}, from
the {\em Casablanca} screenplay in Figure \ref{script-fig}. In the
film, this request is delivered in pleasant tones.

\begin{figure}[htb]
\begin{small}
\begin{center}
\begin{tabular}{|l|}
\hline  
(Laszlo and Ilsa enter Rick's Cafe) \\

Headwaiter: Yes, M'sieur? \\

Laszlo: I reserved a table.  Victor Laszlo. \\

Waiter: Yes, M'sieur Laszlo.  Right this way. \\

(Laszlo and Ilsa follow the waiter to a table) \\

Laszlo: Two cointreaux, please. \\

Waiter: Yes, M'sieur. \\

Laszlo: (to Ilsa) I saw no one of Ugarte's description. \\

Ilsa: Victor, I feel somehow we shouldn't stay here. \\
\hline 
\end{tabular}
\end{center}
\end{small}
\caption{Excerpt from the {\em Casablanca} script.} 
\label{script-fig}
\end{figure} 

\eenumsentence{\item Two cointreaux, please.
\label{las-req-examp}}

However, consider the alternative stylistic realizations in \ex{1}
for requesting two cointreaux:

\eenumsentence{
\item Bring us two cointreaux, right away. 
\item You must bring us two cointreaux.  
\item We don't have two cointreaux, yet. 
\item You wouldn't want to bring us two cointreaux, would you?
}
\label{coint-choice-examps} 

Clearly, speakers make stylistic choices when they realize their
communicative intentions, and their realizations express their
character and personality. And, based on these stylistic
realizations, listeners draw inferences about the character and the
personality of the speaker. Thus, algorithms for LSI are important
for any domain in which agents speak, such as characters for
interactive drama systems, multimodal interface agents and spoken
dialogue
agents\cite{cassell,LoyallBates95,Rich94,Maes94,HR94,Kamm95}.

% THIS is here to get it top of second page
\begin{figure*}[t]
\begin{center} \rule{12cm}{.2mm}
\begin{tabular}{ll} 
header: & {\sc request-act}(speaker, hearer, action)\\
precondition:& {\sc want}(speaker,action) \\
             & {\sc cando}(hearer,action) \\
decomposition-1:& surface-request(speaker,hearer,action) \\
decomposition-2:& surface-request(speaker,hearer, {\sc informif} 
(hearer,speaker, {\sc cando}(hearer,action))) \\
decomposition-3:& surface-inform(speaker,hearer, $\neg$({\sc 
cando}(speaker,action))) \\
decomposition-4:& surface-inform(speaker,hearer, {\sc
want}(speaker,action)) \\
effects: & {\sc want}(hearer,action) \\
               & {\sc know}(hearer, {\sc want}( speaker, action))\\
constraint: & {\sc agent} (action,hearer) 
\end{tabular}
\newline \rule{12cm}{.2mm}
\caption{Definition of the {\sc request-act} plan operator from
Litman and Allen, 1990} 
\end{center} 
\protect\label{litman-fig} 
\end{figure*}

Our work on LSI draws from two theoretical bases: computational
work on {\sc speech acts}\cite{Allen79,Cohen78,Litman85}, and
social anthropology and linguistics research on social
interaction\cite{Goffman83,BrownLevinson87}. The {\em Speech Acts}
section introduces the components of speech act theory that we draw
on; the {\em Social Interaction and Linguistic Style} section
discusses in detail Brown and Levinson's theory of linguistic
social interaction. We argue that these two theories in combination
yield a rich generative source of different characterizations for
artificial agents. The {\em Computing Linguistic Style} section
then explains how these theories provide the basis for generating
the improvisations such as those in \ex{0}, above. The {\em
Implementing Emotional Dispositions} section discusses how we
augment these improvisations by selecting for the speaker an
emotional disposition and its attendant acoustical
correlates\cite{Cahn90}. The {\em Examples} section illustrates how
the theory is implemented in the domain of interactive story and
dialogue. Finally we discuss how LSI extends and differs from other
recent approaches to both interactive drama and text generation and
propose useful extensions to our current work.

\section{Speech Acts}
\label{sact-sec}

Speech acts were first proposed as a small set of communicative
intentions such as {\sc request} or {\sc inform} that underly all
utterance production\cite{Searle75}. In any language based
application, interactive dialogue can be represented as sequences
of speech acts by multiple characters. Therefore, LSI uses speech
acts as the abstract representation for utterances, and plans as
the basis for improvisation --- each spoken utterance is
represented as an instantiation of a plan operator and these
instantiations are interleaved with descriptions of physical acts
in a real or simulated world.

The inventory of speech acts is defined by the application. Ours
consists of the initiating acts of {\sc inform, offer} and two
types of {\sc request}: {\sc request-info} and {\sc request-act}.
We also use three types of response speech acts for acceptance and
rejection, corresponding to each major type of initiating act: {\sc
accept-inform, accept-offer and accept-request}; and {\sc
reject-inform}, {\sc reject-offer} and {\sc reject-request}.

Each speech act definition includes (a) specifying the conditions
under which a speaker performing the speech act could be successful
at achieving a communicative intention, and (b) specifying the
effects on the hearer if the speaker is successful. Earlier
computational work proposed that speech acts should be implemented
in a standard AI planning system as plan operators that include the
act's {\sc decomposition}, {\sc preconditions} and {\sc effects},
thereby enabling computer agents to plan utterances in the same way
that they plan physical acts\cite{Allen79,Cohen78,Litman85}. An
example plan-based representation of a {\sc request-act} (for
example, Laszlo's request in \ref{las-req-examp}) based on Litman
and Allen's work, is given in Figure \ref{litman-fig}\cite{LA90}.

A critical basis of our improvisation algorithms is speech act
theory's distinction between the underlying intention of a speech
act, and the surface forms of the utterance that can {\sc realize}
the speech act. This distinction is seen in Figure
\ref{litman-fig}: the {\sc request-act} speech act specifies an
underlying intention (the desired effect) of the speaker getting
the hearer to do (or want to do) a particular action; while the
four decompositions specify the different ways that the underlying
speech act can be realized by {\em surface} speech acts, that is,
by particular sentential forms such as declarative sentences or
questions. For example, the sentential equivalents of
decompositions 1 to 4 in Figure \ref{litman-fig} might be those in
\ex{1}a to \ex{1}d respectively, where {\it action} represents an
action description:

\eenumsentence{
\item  Do {\it action}.
\item  Can you do {\it action}?
\item  I can't do {\it action}.
\item  I want {\it action}.
}

Our algorithms for improvisation, to be discussed in the {\em
Computing Linguistic Style} section, are mechanisms for deciding
how to realize a given underlying intention as a particular surface
form. While previous work on dialogue generation has focused on
informational motivations and effects\cite{MooreParis93}, we focus
here on the impact of social and affective parameters on the
selection of utterance form and content.

\section{Social Interaction and Linguistic Style}
\label{bl-sec}

Whenever agents realize a particular speech act, they make choices
about the linguistic style with which that act is realized. Our
main idea is that all these choices have a major effect on our
perception of an agent's character and personality. Given the goal
of achieving a particular communicative intention in a given social
setting, an agent must choose among all the possible variations in
{\sc semantic content}, {\sc syntactic form} and {\sc acoustical
realization}. We call these choices a {\sc strategy} for realizing
a particular communicative intention.

The generative account we present is derived from Brown and
Levinson's theory of social interaction\cite{BrownLevinson87} in
which they identify a number of different variables and give
examples of how different values for the variables produce
different communicative outcomes. In LSI, we take their framework,
refine its specification where necessary, and specify the
computational mechanisms required to implement it.\footnote{Due to
space constraints, we are unable here to present a full exegesis of
their theory, the interested reader is referred to
\cite{BrownLevinson87}.}

\subsubsection{Maintaining public face}

An important basis of the theory is that all agents have and know each
other to have:

\begin{enumerate} 

\item {\sc Face}: An agent's public self image, which consists of the
desire for:

\begin{enumerate}
\item {\sc Autonomy}: Freedom of action and freedom from
imposition by other agents; 
\item {\sc Approval}: A positive consistent self-image or
personality that is appreciated and approved of by other agents;
\end{enumerate}

\item Capabilities for {\sc rational reasoning} such as means-end
reasoning, deliberation, and plan recognition.
\end{enumerate}

\subsubsection{Social variables and face}

Given the desire to maintain their own and others' face, and beliefs
about their own and others' rationality, the agents' algorithm for
choosing a strategy for realizing a particular speech act relies on
evaluating three socially determined variables:

\begin{enumerate} 
\item D(S,H): the {\sc social distance} between the speaker and hearer.
\item P(H,S): the {\sc power} that the hearer has over the speaker.
\item R$_\alpha$: a {\sc ranking of imposition} for the act $\alpha$ 
under discussion.
\end{enumerate}

Human agents use personal experience, background knowledge, and
cultural norms to determine the values for these variables. For
example, {\sc social distance} often depends on how well S and H
know one another, but also on social class and status. {\sc Power}
comes from many sources, but often arises from the ability of S to
control access to goods that H wants, such as money.

The {\sc ranking of imposition} relies on the fact that all agents'
basic desires include the desire for autonomy and approval. Thus
particular speech act types can be ranked as higher impositions
simply by how they relate to agents' basic desires. 

Speech acts that can function as a threat to H's desire for
autonomy include those that predicate some future act of H, as well
as speech acts that predicate some future act of S toward H, such
as offers, which put pressure on H to accept or reject them. This
means that the act types of {\sc request-inform}, {\sc request-act}
and {\sc offer} threaten H's desire for autonomy. The {\sc inform}
speech act also threatens H's desire for autonomy on the basis that
it is an attempt by S to affect H's mental state.

Speech acts that threaten H's desire for approval include all
rejections, including the act types {\sc reject-inform}, {\sc
reject-offer} and {\sc reject-request}.\footnote{ Other speech acts
not in our inventory, such as criticisms and complaints, also
threaten H's desire for approval\cite{BrownLevinson87}.}

Given our inventory of speech acts, and the range of the variables
D and P, we instantiate the theory with the ranking of imposition
R$_\alpha$ based on the speech act type, as shown in Figure
\ref{rx-fig} below.\footnote{The values we use here serve to
illustrate the model and range of phenomena. The actual values of
the ranking of imposition need to be empirically determined with
respect to the culture being modeled. We also discuss in our
concluding section how R$_\alpha$ should be a function of both
speech act type and propositional content, rather than purely
speech act type as we do here.}

\begin{small}
\begin{figure}[htb]
\begin{center} 
\begin{tabular}{|l|c|}
 \hline
Speech Act & R$_\alpha$  \\
 \hline \hline
accept-request & 5 \\ 
\hline 
accept-inform& 5\\ 
\hline 
accept-offer& 10 \\
\hline
inform & 15 \\
\hline 
request-info & 20 \\
\hline 
offer & 25 \\
\hline 
reject-offer &30 \\ 
\hline 
reject-inform &35\\
\hline 
reject-request &40 \\ 
\hline 
request-act & 45 \\ 
\hline
\end{tabular}     
\caption{A ranking R$_\alpha$ on imposition of various types of
speech acts with values from 1 to 50. }
\end{center} 
\label{rx-fig}
\end{figure}
\end{small}

\subsubsection{Linguistic style strategies and social variables}

As social and rational actors, S and H attempt to avoid threats to
one another's face. Given values for social distance D(S,H), power
P(H,S) and ranking of imposition R$_\alpha$, the agent S estimates
the {\sc threat} $\Theta$ to H of performing the speech act
$\alpha$ by simply summing these variables as in equation \ex{1}:

\enumsentence{$\Theta$ $=$ D(S,H) + P(H,S) + R$_\alpha$
\label{calc-weight-eq}}

Once a value for $\Theta$ has been calculated, the agent uses it to
choose among one of the following four strategies for executing a
speech act:\footnote{Brown and Levinson include a strategy of not
executing the speech act at all because the face threat is too
great.}

\eenumsentence{
\item {\sc Direct}: Do the act directly. 
\item {\sc Approval-Oriented}: Orient the realization of the act to
H's desire for approval;
\item {\sc Autonomy-Oriented}: Orient the realization of the act
to H's desire for autonomy;
\item {\sc Off-Record}: Do the act off record by hinting, and/or by
ensuring that the interpretation of the utterance is
ambiguous. 
\label{strat-choices}}

The lowest values of $\Theta$ lead to the {\sc direct} strategy and
higher values lead to the {\sc off-record} strategy. In LSI, the
range for each of the social variables D, P and R$_\alpha$ is
between 0 and 50. Therefore, the $\Theta$ sum will range from 0 to
150. {\sc direct} strategies correspond to $\Theta$ values through
50, {\sc approval-oriented} strategies to $\Theta$ values from 51
to 80, {\sc autonomy-oriented} strategies for $\Theta$ values from
81 to 120 and {\sc off-record} strategies for $\Theta$ values from
121 to 150.\footnote{Again these values are estimates selected for
illustrative purposes.}

Each strategy can be realized by a wide range of sub-strategies,
whose {\sc semantic content} is selected from the plan-based
representation for a speech act and whose {\sc syntactic form} is
selected from a library of syntactic forms. And since there are
many ways to realize each strategy, realizations within particular
ranges are heuristically assigned to the upper or lower end of the
scale, or assigned to the same values of the scale to support
random variation.

\subsection{Emotion as an element of linguistic style}

Varying the affect of the spoken realization is a critical aspect
of linguistic style. Although Brown and Levinson state that
expressions of strong emotion threaten both S and H's desires for
approval and autonomy, they do not further specify the relation
between strategies for selecting {\sc semantic content} and {\sc
syntactic form}, and those for selecting the {\sc acoustical
realizations} in the utterance which most directly express
emotions. In order to explore this interaction, we adopt a very
simple view of emotional expression: emotional disposition is an
orthogonal dimension to social variables, and each character is
simply assigned an emotional disposition at the start.

\section{Computing Linguistic Style}
\label{improv-sec}

\begin{figure*}[t]
\centerline{\psfig{figure=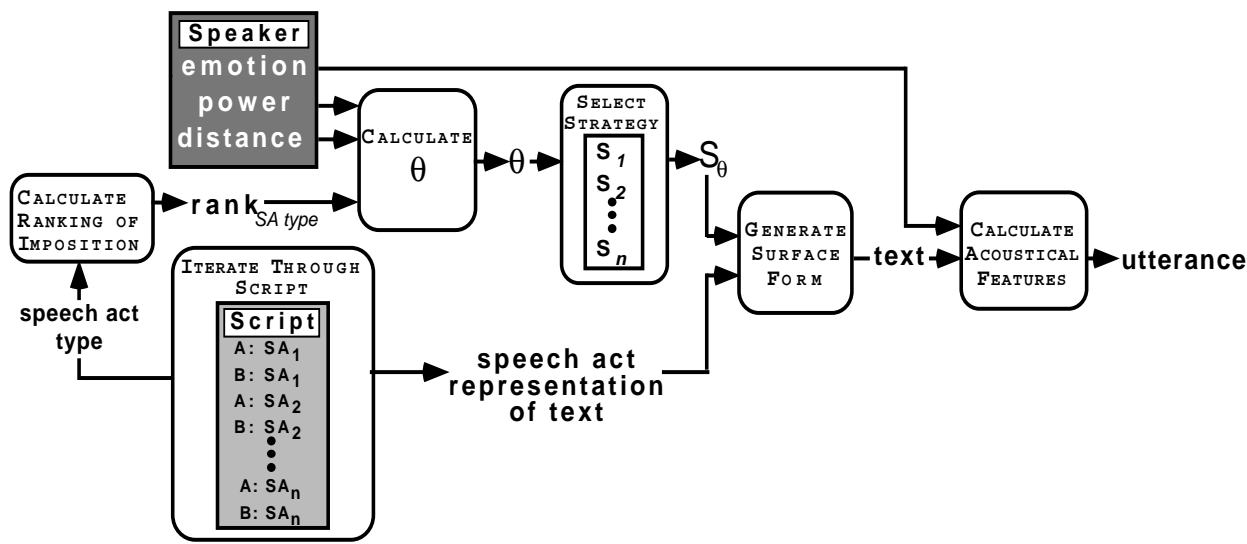,height=2.8in,width=6.5in}}
\caption{Overview of LSI Algorithm}
\label{lsi-fig}
\end{figure*}

Because LSI is defined on the basis of speech act types alone, what
we have described so far is domain independent. However, the
content of each speech act is domain specific. For example, in
Figure \ref{litman-fig}, domain specific contain is represented by
the {\tt action} variable in the definition of {\sc request-act}.
Similarly, the domain specific content in an {\sc inform} speech
act is represented by a {\tt proposition} variable. Thus to test
LSI, specific domains must be represented in terms of the actions
and propositions of that domain. For example, Figure
\ref{cointreaux-fig} represents the domain specific action of
serving two cointreaux.

We have tested LSI on speech acts derived from two domains: a
task-oriented dialogue in which two agents discuss furnishing a two
room house\cite{Walker96a}, and a segment of the {\em Casablanca}
script shown in Figure
\ref{script-fig}\cite{viva,spline}.\footnote{The task oriented
dialogue representation is generated off-line by a planner, while
the {\em Casablanca} script speech act representation is
constructed by hand. In both cases, we use the generator
FUF\cite{Elhadad92} to generate surface forms. Because FUF does not
operate directly on predicate logic representations used in plans,
we therefore augment these with manually generated FUF equivalents.
Future implementations will include a transducer that generates FUF
forms automatically from plan representations.}

\begin{figure*}[t]\begin{center} 

\rule{7cm}{.2mm}

\begin{tabular}{ll} header: & {\sc serve}(waiter, customer, two-cointreaux)\\

precondition:& {\sc has}(restaurant, two-cointreaux) \\

decomposition:& {\sc bring}(waiter, customer, two-cointreaux)\\

effects: & {\sc has}(customer, two-cointreaux)
\end{tabular}

\rule{7cm}{.2mm}
\caption{A possible plan in the restaurant domain for serving two cointreaux} 
\end{center} 
\label{cointreaux-fig} 
\end{figure*}

As shown in Figure \ref{lsi-fig}, LSI takes an input a sequence of
speech acts representing a dialogue, and a {\sc social structure}
which consists of a value between 0 and 50, for both social
distance D and for power P, for each pair of agents in the
dialogue. Then, for each speech act in the script or the dialogue,
the speaker determines the social distance D between him/herself
and the hearer, the power P that the hearer has over him/her, and
the value on R$_\alpha$ for the speech act type as in Figure
\ref{rx-fig}. Then by equation \ref{calc-weight-eq}, the speaker
calculates the value of $\Theta$, and uses this to select one of
the strategies given in 5 above. %\ref{strat-choices}.

We will now demonstrate how the algorithm operates, by showing how
different linguistic strategies result from different social
structures. In each case we will use the example from {\em
Casablanca}, in which Laszlo orders two cointreaux from Emil, and
assume that the algorithm operates on the representations in
Figures \ref{litman-fig} and
\ref{cointreaux-fig}.\footnote{Actually we will derive some of the
decompositions in Litman's definition by
rule\cite{AP80,GordonLakoff71}.} Since there are many more
realizations of the strategies than can be discussed here,
interested readers are referred to \cite{BrownLevinson87}.

\subsection{Direct strategies}

Direct strategies result from social structures in which both
social distance D and power P are small. In the case of our two
cointreaux example, imagine that Laszlo and Emil are old friends,
and that Emil, as the waiter, has no power over Laszlo. This could
be modeled in our framework with a social structure in which the
social distance D between Emil and Laszlo is 4 and the power P that
Emil has over Laszlo is 0. According to Figure \ref{rx-fig}, the
R$_\alpha$ for {\sc request-act} is 45. Using equation 4 and the
values for P, D and R$_\alpha$, the value for $\Theta$ is 49,
leading Laszlo to select a direct form strategy for realizing his
request.

The realizations for all {\sc direct} forms, irrespective of speech
act type, are based on the {\sc semantic content} of the
decomposition step of the speech act. Each speech act type has an
associated {\em default} {\sc syntactic form}. 

For example, in the case of {\sc request-acts} we assume that the
default syntactic form is an imperative.\footnote{For speech acts
such as {\sc inform}, the default syntactic form is a declarative
sentence, and for speech acts which are subtypes of {\sc accept} or
{\sc reject}, the default forms are {\it Okay, Yes} or {\it No},
respectively.} Thus the simplest strategy for realizing a direct
form is the {\sc realize-direct-strategy}: Realize the content of
the decomposition step with its associated default syntactic form.
For a request such as {\it Two cointreaux, please}, this would
result in an utterance such as: 

\enumsentence{Bring us two cointreaux.}

Direct realizations can also be ordered within the range of 0 to 50
so that lower values correspond to styles that convey that H has no
power (P is low). One way to make a {\sc request-act} is the {\sc
power-direct-strategy}: Add {\it you must} or {\it right away} to
the direct form. This is illustrated in \ex{1} and \ex{2}:

\enumsentence{Bring us two cointreaux right away.}

\enumsentence{You must bring us two cointreaux.}

\subsection{Approval oriented strategies}

Approval oriented strategies result from social structures in which
there are minor differences in both power P and social distance D
between the interactants, so that these factors play a weak role in
strategy selection. Strategies for orienting the realization of a
speech act to the hearer's desire for approval include intensifying
interest or attention to H, implying that S and H are cooperators
who have the same perspective or desires, or conveying that S and H
are part of the same social group or are friends.

One way to convey that S and H have the same desires when making a
request is the {\sc optimism-approval-strategy}: S expresses
optimism that H will want to do what S wants H to do. This strategy
results from selecting the semantic content to be realized from the
{\it want hearer action} effect of the request-act (as in Figure
\ref{litman-fig})\footnote{A similar strategy of assuming that the
effect already holds can also be used for {\sc inform} speech
acts.}, and realizing this semantic content with a declarative
sentence that includes a tag question. This strategy results in
surface forms such as:

\enumsentence{You'd like to bring us two cointreaux, wouldn't you?}

One way to imply that S and H are in the same social group and that
S believes that the relative P between himself and H is small is
the {\sc group-approval-strategy}: Use in-group address forms such
as {\it buddy, mate, honey, doll, my man}, depending on the group.
For a request, this is implemented by concatenating an in-group
address form, {\it my man}, to the direct realization of the speech
act, resulting in surface forms such as:

\enumsentence{Hey Emil, my man, bring us two cointreaux.}

For {\sc accept-offer} or {\sc accept-request} speech acts,
approval oriented forms are those that explicitly assert the {\sc
want} effect of the offer or request speech act, such as:

\enumsentence{I'd be glad to.}

and 
\enumsentence{With pleasure.}

For rejections, approval oriented forms are those by which H
affirms a social relationship with S such as: 

\enumsentence{I'm sorry, I can't. Normally I'd love to.}

\subsection{Autonomy oriented strategies}

Autonomy oriented strategies result from social structures in which
there are significant differences between the two agents in either
power P or social distance D. Under these circumstances S will
choose strategies that make minimal assumptions about H's wants and
desires, leaving H the option not to do the act, and disassociate S
from possible infringement of H's autonomy.

Note that the effect field in Figure \ref{litman-fig} encodes
information about H's wants and desires. Thus, one rule is to be
pessimistic about H's desires. This can be achieved by selecting
semantic content from this effect field with the {\sc
negate-effect-autonomy-strategy}: State that the want effect
doesn't hold. This produces a form such as:

\enumsentence{You wouldn't want to bring us two cointreaux would you?}

In addition, note that the precondition field in Figure
\ref{litman-fig} encodes information about H's abilities. One way
of leaving H the option not to do the act is for S to produce a
query with this precondition as the semantic content, leaving H the
option of saying that s/he is unable to do the act. This is the
{\sc query-ability-autonomy-strategy}, which results in forms such
as:

\enumsentence{Can you bring us two cointreaux?}

One way of disassociating S and H from an autonomy infringement is
to produce an indirect form of a request with the {\sc
assert-want-precondition-autonomy-strategy}: State that the want
precondition holds. This results in forms such as:

\enumsentence{We'd like two cointreaux.}

Another strategy for avoiding an autonomy infringement is the {\sc
impersonalize-actor-autonomy-strategy}: Impersonalize who actually
performs the requested act. This results in proposals with no actor
specified. It is also possible to produce proposals in which the
act itself is unspecified, by selecting the semantic content for
the request from the effect field of the domain act. For example,
in Figure \ref{cointreaux-fig}, the effect is that the customer has
two cointreaux. Using this field as the semantic content results in
surface forms such as:

\enumsentence{Let us have two cointreaux.}

{\sc inform} speech acts also have realizations that are autonomy
oriented. An {\sc inform} speech act can impinge on H's autonomy
concerning what s/he wants to believe. One way to orient to H's
autonomy is to soften the strength of an assertion by {\sc hedging}
it \cite{Prince80}. For example, consider Laszlo's utterance of
{\it I reserved a table}. This can be hedged by simply embedding
the declarative sentence, which is produced from the decomposition
step of the plan for an {\sc inform}, with hedging phrases such as
{\it I feel, I believe, It seems, As you may know, I think, I
heard}, or adding other hedges such as {\it somehow, sort of, kind
of} to the verb phrase. This strategy is encapsulated in \ex{1} and
produces forms such as \ex{2}:

\enumsentence{{\sc hedge-inform-strategy}: Augment any inform
statement with either a pre-sentential or a verbal hedge.}

\enumsentence{I believe I reserved a table.}

An example of hedging in the original script (Figure \ref{script-fig})
is Ilsa's assertion:

\enumsentence{Victor, I feel somehow we shouldn't stay here.}

Hedging the strength of the assertion can also function as an
approval oriented strategy since it is a simple way to avoid
disagreement. 

\subsection{Off record strategies}

Off record strategies result from social situations in which there
are significant values for social distance D or major discrepancies
in power P between two agents, or from an act that is a large
imposition on H. Tactics for going off record are difficult to
implement because strategies for doing so involve indirect
inference paths that are difficult to model computationally. There
are, however, several simple ways to make a request off record by
constructing hints from plan-based representations. One strategy is
the {\sc assert-negation-domain-effect-strategy}, in which S
asserts that the effect of the domain plan does not hold, as in:

\enumsentence{We don't have two cointreaux yet.}

Another strategy is the {\sc
assert-domain-precondition-holds-strategy:} Assert that the
precondition of the domain plan holds. For example, Laszlo's
utterance of {\it I reserved a table} is a statement that the
domain precondition for being shown to a table holds. Thus the
original realization in the script is an off record form.

Another strategy is the {\sc
abstract-agent-and-negate-effect-strategy:} Select the semantic
content as the decomposition of the domain plan, abstract the agent
role, and negate the assertion of the decomposition. This leads to
an implicature\cite{Hirschberg85}. The result is shown below:

\enumsentence{Someone hasn't brought us two cointreaux.}

In the current implementation of LSI, autonomy oriented forms are
sometimes substituted for off record forms in order to provide more
variability when characters choose to go off record.

\section{Implementing Emotional Dispositions}
\label{affect-sec}

Once a character's emotional disposition has been set, all of that
character's utterances are synthesized with the acoustical
correlates of that emotion. We implement this by drawing on Cahn's
theory of expressing affect in synthesized speech\cite{Cahn90}, and
use a version of her Affect Editor program developed expressly for
interactive theater and simulated conversation. 

The Affect Editor computes instructions for a speech synthesizer
(so far, the DECtalk3 and 4.1) so that it produces emotional and
expressive synthesized speech. The output is a set of synthesizer
instructions; the input is a combination of text and acoustical
parameter values. The parameters (seventeen in all) control the
presence in the speech signal of various aspects of pitch, timing,
voice quality and phoneme quality. 

Because some of the acoustical properties are moderated by
linguistic properties of the text, the words in the text must be
annotated for part of speech, focus information (expressed as a
likelihood of receiving intonational stress, that is, as the
inverse of the accessibility of items in memory), and then the text
itself marked with all possible phrase boundaries according to
syntax and grammatical role. 

The acoustical parameters have numerical values. Their adjustment
around zero --- representing neutral affect --- allows various
shadings of emotional expression, for example, from calm to sad to
completely dejected, or from enthusiasm to harsh anger. Our current
LSI implementations make use of parameter value sets for seven
emotional dispositions: Angry, Annoyed, Disgusted, Distraught,
Gruff, Pleasant and Sad.

\begin{figure*}[htb]
%\begin{small}
\begin{center}
\begin{tabular}{|lc|}
\hline 
(Laszlo and Ilsa enter Rick's Cafe) & \\

Headwaiter: Yes, M'sieur? & ({\sc offer}) \\

Laszlo: I reserved a table. Victor Laszlo. & ({\sc request-act}) \\

Waiter: Yes, M'sieur Laszlo. Right this way. & ({\sc
accept-request}) \\

(Laszlo and Ilsa follow the waiter to a table) & \\

Laszlo: Two cointreaux, please. & ({\sc request-act}) \\

Waiter: Yes, M'sieur. & ({\sc accept-request})\\ 
\hline
\end{tabular}
\end{center}
%\end{small}
\caption{Assumed Speech Acts for an excerpt from the {\em Casablanca} 
script.} 
\label{script-fig2}
\end{figure*} 

\section{Example Runs of Linguistic Style Improvisation}
\label{res-sec}

To demonstrate the effect of LSI, we apply it to the first five
lines of the {\em Casablanca} script in Figure \ref{script-fig},
where agent A is Laszlo and agent B is the waiter. We provide an
underlying abstract representation for this excerpt in terms of
speech acts as specified in Figure \ref{script-fig2}. We use
extreme power and social distance parameter settings in the
examples to demonstrate the range of variation that is possible.

\subsubsection {A direct/angry speaker with an
approval-oriented/pleasant hearer} 

In a social structure in which A's emotional disposition is angry,
and B's is pleasant, modeled by setting D(A,B) $=$ 0, P(B,A) $=$ 0,
D(B,A) $=$ 30, and P(A,B) $=$ 30, A will choose direct strategies
and an angry delivery, and B will choose approval oriented
strategies, delivered in pleasant tones. The result of this social
structure applied to the {\em Casablanca} excerpt is:

%Laszlo: direct+power, angry
%Waiter: approval, pleasant}
\eenumsentence{
\item[W:]Could I help you?
\item[L:]You must take us to a table. I am Victor Laszlo.
\item[W:]It's a pleasure.
\item[L:]Bring us two cointreaux, right away
\item[W:]I'd be glad to.
}

\subsubsection{An autonomy-oriented/distraught speaker with a
direct/pleasant hearer}

In a social structure where A's emotional disposition is
distraught, and B's is pleasant, modeled by setting D(A,B) $=$ 40,
P(B,A) $=$ 40, D(B,A) $=$ 0, and P(A,B) $=$ 0, A will choose
autonomy oriented strategies and a distraught delivery. and B will
choose the lower end of direct strategies and a pleasant delivery.
The effect of this social structure on the {\em Casablanca} excerpt
is:

%%Laszlo: autonomy, distraught
%%Waiter: direct+power, pleasant
\eenumsentence{
\item[W:]I will help you
\item[L:]Can you take us to a table?
As you may know, I am Victor Laszlo
\item[W:]Yes, if you insist.
\item[L:]You wouldn't want to bring us two cointreaux, would you?
\item[W:]Yes, if I must.
}

The values that produce \ex{0} portray Laszlo as a wimp, for
several reasons. First, Laszlo, who is the customer, is orienting
to the waiter's autonomy. Second, the distraught delivery is very
high pitched and tentative. Finally, the fact that the waiter is
rude highlights their differences in linguistic style.

\section{Related Work}

There are two areas of related work: recent work on interactive
drama systems ---in particular, Hayes-Roth's work on improvisation
by computer characters; and the longer running body of work on
natural language generation.

\subsubsection{Interactive drama systems} 

In empirical studies of human reactions to lifelike computer
characters, Nass {\it et al.}\cite{NR} show that linguistic style
leads to specific inferences about character. However, they rely on
pre-scripted linguistic forms to demonstrate its effects and no
generative mechanism is supplied. Other work in this area, for
example, that of Maes {\it et al.}\cite{Maes94} and Loyall and
Bates \cite{LoyallBates95} has focused on the behavior on
non-speaking animals, so that linguistic style has not been
considered. Where characters do speak, their utterances are in the
main pre-scripted \cite{GeneBall}, or generation does not focus on
variations in linguistic style\cite{cassell}.

Hayes-Roth's work on improvisation does allow for linguistic
variation, but this arises by selection from a finite set of forms,
and again no generative mechanism is given\cite{HR94,HR95}. However
this work provides a useful set of requirements for improvisation
mechanisms of computer characters\cite{HR95}, which our mechanisms
for LSI satisfy:

\begin{enumerate}
\item {\it Interesting variability} in a character's interpretation
of a given direction on different occasions;
\item {\it Random variability} in the way a character performs a
specific behavior on different occasions;
\item {\it Idiosyncrasies} in the behaviors of different characters;
\item {\it Plausible motivations} for character's behavior;
\item {\it Recognizable emotions} associated with character's behaviors
and interactions.
\end{enumerate}
 
The dialogues in 23 and 24 demonstrate that social structure
variables produce {\it interesting variability}, {\it random
variability}, and {\it idiosyncrasies}. In addition, because Brown
and Levinson's theory is based on empirical observation of human
interaction in many cultures, a theory of LSI based on it satisfies
Hayes-Roth's last two criteria. Since the theory captures
linguistic universals, human users should be able to ascribe
plausible motivations and recognize the emotions associated with a
character's behavior. Especially, the motivations the theory
ascribes are not only descriptive and explanatory, but {\em
predictive} and {\em generative}.

\subsubsection{Text generation}

Previous work on natural language generation has addressed the
problems of how surface forms can be generated from underlying
speech acts \cite{MooreParis93,Cohen78,Dalethesis}, {\it inter
alia}. However in the main, the variables that determine linguistic
choice have all been task-related. The generation research has
therefore addressed the role of linguistic choice in indicating
information structure; foregrounding and backgrounding information;
reducing cognitive overload, and the impact of these factors on
inducing change in the hearer's beliefs. This task oriented
perspective ignores other aspects of choice and interaction,
namely, agents' motivations, and socially appropriate responses and
behavior.

One exception is the work of Hovy\cite{Hovy93}, who does consider
the effect of social factors on generation. However, Hovy is
concerned with generating news stories (text) which, in speech act
terms are sequences of {\sc inform} speech acts. In contrast, our
work focuses on the generation of conversation, which requires a
much wider range of speech acts. Furthermore, the news story genre
affords fewer opportunities for social factors to affect generation
given the anonymity of the generic text reader.

\section{Discussion and Future Work}
\label{conc-sec}

In this paper, we have argued that linguistic style is an
under-researched aspect of character, and presented a theory of,
and algorithms for, Linguistic Style Improvisation by computer
characters. This work expands the set of parameters that have been
investigated in research on natural language generation of
conversational speech.

Possible interesting extensions to our work would be to introduce
social feedback into our model, allowing linguistic actions to
directly affect the {\sc social structure} in the course of an
interaction. We hope to explore a reciprocal feedback loop to
social structure, in which, for example, one agent's linguistic
friendliness results in another agent adjusting their beliefs about
social distance, and hence changing the second agent's future
linguistic strategies. This should result in interpretable and
interesting changes in the way two agents treat one another over
the course of a social interaction. We also hope to examine in more
detail the relationship of acoustical expression of emotions to
choices about linguistic semantic content and syntactic form.

Another possible extension concerns a more complex function for
calculating the {\sc ranking of imposition} R$_\alpha$. The problem
is that R$_\alpha$ should be a function of {\em both} the speech
act type, and the type of the action {\em in the domain}. For
example, a {\sc request-act} that H pass the salt is less of an
imposition than a {\sc request-act} that H give S five dollars. We
conjecture that a function for R$_\alpha$ could be based on inputs
$\alpha$ and a domain act $\delta$, if the speech act planner could
access information about the effort involved with the execution of
the domain act $\delta$.

In sum, we have shown how LSI can be applied to computer characters
in both interactive fiction and task-oriented dialogue simulation.
In future work, we hope to investigate applying the same mechanisms
to characters for personal assistants for spoken language
interfaces\cite{GeneBall,Kamm95,Yankelovich95}. We believe that the
combination of dimensions we have focused on provides a motivated
and artistically interesting basis for making choices about
linguistic style, that these choices are closely related to human
perceptions of character and personality, and that they provide a
rich generative source of linguistic behaviors for lifelike
computer characters.

\section{Acknowledgments}

Thanks to Gene Ball, Justine Cassell, Larry Friedlander, Dawn
Griesbach, Don Martinelli, Phil Stenton, and Loren Terveen for
interesting discussions about linguistic style, to Charles Callaway
and Michael Elhadad for consultation on using FUF, to Obed Torres
for implementation work on VIVA, and to Wendy Plesniak for artistic
inspiration.

\newcommand{\etalchar}[1]{$^{#1}$}

%\bibliography{acm}

\end{document}